\documentclass[1p]{elsarticle}

\usepackage[utf8]{inputenc} 
\usepackage[T1]{fontenc}    
\usepackage{hyperref}       
\usepackage{url}            
\usepackage{booktabs}       
\usepackage{amsfonts}       
\usepackage{nicefrac}       
\usepackage{microtype}      
\usepackage{xcolor}         
\usepackage{amsmath}
\usepackage{multirow}
\usepackage{booktabs}
\usepackage{graphicx}
\usepackage{algorithm}
\usepackage{algpseudocode}
\usepackage{subfigure}

\newcommand{\yv}{\mathbf{y}}
\newcommand{\Xv}{\mathbf{X}}

\title{Interpretable mixture of experts for time series prediction under recurrent and non-recurrent conditions}

\author[1]{Zemian Ke}
\author[1]{Haocheng Duan}
\author[1,2]{Sean Qian\corref{cor1}}
\cortext[cor1]{Corresponding author, seanqian@cmu.edu}

\affiliation[1]{organization={Department of Civil and Environmental Engineering, Carnegie Mellon University},
city={Pittsburgh},
country={USA}}
\affiliation[2]{organization={Heinz College, Carnegie Mellon University},
city={Pittsburgh},
country={USA}}

\begin{document}

\begin{abstract}
Non-recurrent conditions caused by incidents are different from recurrent conditions that follow periodic patterns. Existing traffic speed prediction studies are incident-agnostic and use one single model to learn all possible patterns from these drastically diverse conditions. This study proposes a novel Mixture of Experts (MoE) model to improve traffic speed prediction under two separate conditions, recurrent and non-recurrent (i.e., with and without incidents). The MoE leverages separate recurrent and non-recurrent expert models (Temporal Fusion Transformers) to capture the distinct patterns of each traffic condition. Additionally, we propose a training pipeline for non-recurrent models to remedy the limited data issues. To train our model, multi-source datasets, including traffic speed, incident reports, and weather data, are integrated and processed to be informative features. Evaluations on a real road network demonstrate that the MoE achieves lower errors compared to other benchmark algorithms. The model predictions are interpreted in terms of temporal dependences and variable importances in each condition separately to shed light on the differences between recurrent and non-recurrent conditions.
\end{abstract}

\maketitle
\section{Introduction}
Transportation networks, designed to serve people with efficient mobility, are plagued by congestion. In 2019, U.S. roadways witnessed 8.8 billion hours of travel delay and 55 hours of delay per driving commuter \cite{lasley20212021,hampshire2022pocket}, around half of which is with incidents \cite{fhwa2019} (25\% by accidents, 15\% by weather and 10\% by work zones). Those planned and unplanned incidents (e.g. hazardous weather conditions, accidents, local events, etc.) on the highway networks can catastrophically impact mobility and safety. Mitigating incident impacts requires accurate and ahead-of-curve real-time prediction and proactive operational management \cite{ke2020enhancing, ke2024real}. This study seeks to address the primary challenge of accurately predicting network traffic states. 

Traffic conditions with the occurrence of incidents are regarded as non-recurrent, which are distinct from the recurrent conditions that follow periodic patterns. Non-recurrent conditions caused by incidents are volatile because incidents typically lead to sudden disruptions in the flow of traffic. These incidents create irregular traffic patterns and can significantly impact travel times and congestion levels. In contrast, recurrent conditions are the result of consistent and predictable factors, such as daily rush hours or regular bottlenecks, where traffic patterns follow a routine cycle.

\begin{figure}
    \centering
    \includegraphics[width=0.8\textwidth]{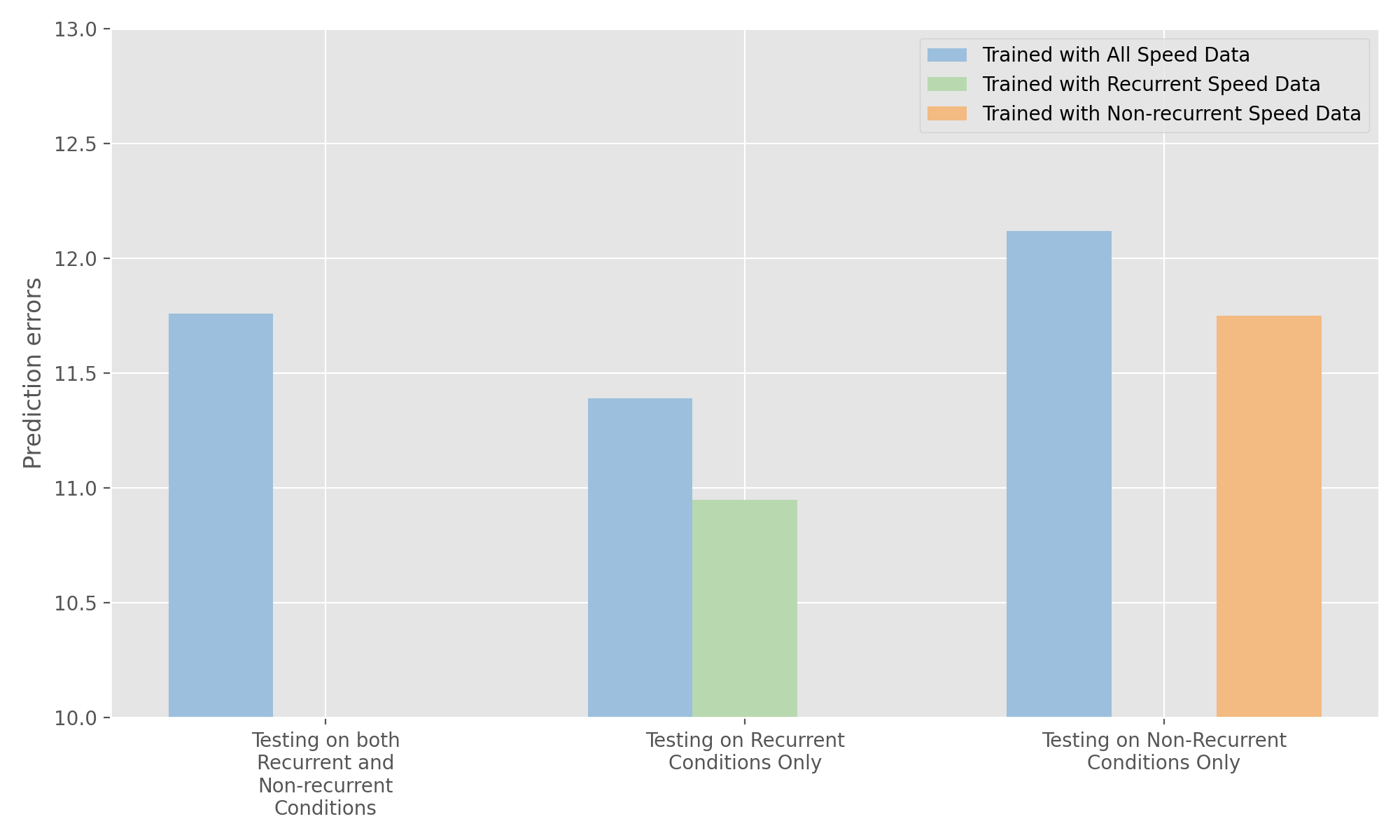}
    \caption{An illustrative example of prediction errors.}
    \label{fig: demo}
\end{figure}

Despite consideration of road incidents (accidents, severe weather impact, etc) in modeling, existing approaches of traffic prediction fail to distinguish between recurrent conditions and non-recurrent conditions. A general consensus is that existing real-time traffic prediction works reasonably well for recurrent traffic, but not yet well for non-recurrent traffic (see Figure \ref{fig: demo} in which a state-of-the-art time series forecasting model is applied). It is less studied to accurately predict traffic states in non-recurrent conditions by learning from multi-source emerging traffic data. What makes it particularly challenging is to work with an incident that occurs at a location/time where incident data is very rare or even nonexistent. It is also unclear how predictions in non-recurrent conditions differ from recurrent ones.

A Mixture of Experts (MoE) \cite{masoudnia2014mixture} is a universal model structure that combines multiple specialized sub-models, or "experts," each tailored to handle different aspects or patterns within a dataset. A gating network determines which expert(s) to utilize for a given input, optimizing performance by leveraging the strengths of each expert for specific types of data or tasks. Inspired by the diverse capabilities of MoE shown in recent large language models (e.g., \cite{jiang2024mixtral}) and the observation that data of recurrent and non-recurrent conditions is heterogeneous, we propose a MoE-based model that contains a recurrent expert model and a non-recurrent expert model. For the expert model backbone, Temporal Fusion Transformer (TFT) \cite{lim2021temporal} is selected because of its superior forecasting capability and interoperability. Moreover, a general training pipeline is proposed for the non-recurrent expert model to remedy the limited data issues.

The contributions of this study can be summarized as follows.
\begin{itemize}
    \item Differentiating from the literature that is incident-agnostic, this study explicitly considers incident occurrences, so as to improve the prediction in both recurrent and non-recurrent conditions.
    \item We propose a MoE model that adopts TFT as expert sub-models and multi-source features are utilized.
    \item We propose a general training pipeline for non-recurrent traffic prediction models to overcome the challenge of extremely limited non-recurrent datasets.
    \item Model predictions are interpreted to shed light on differences of forecasting under recurrent and non-recurrent conditions.
\end{itemize}

The remainder of this paper is organized as follows. Related work is reviewed and summarized in Section \ref{sec: related work}. Section \ref{sec: dataset} introduces the multi-source datasets. Subsequently, section \ref{sec: method} elaborates on data pipeline, model structure, and model training. Experiments and results are included in section \ref{sec: experiments}. Last, the paper concludes with a summary of the main findings in Section \ref{sec: conclusion}.

\section{Related work}
\label{sec: related work}

Leveraging machine learning to predict traffic flow or traffic time has been widely studied in the last decades. First, linear time series analysis has been widely recognized and used for traffic flow/speed prediction, such as Auto-Regressive Intergrated Moving Average (ARIMA) utilized by \cite{pace1998spatiotemporal}, \cite{kamarianakis2005space}, \cite{kamarianakis2003forecasting}, for instance. \cite{guo2014adaptive} adopts Kalman filtering for traffic flow forecasting. Non-parametric regression model was also utilized in \cite{smith2002comparison} and \cite{rahmani2015non}. Classical machine learning models, such as Support Vector machine model, is also used to predict traffic flow \cite{cong2016traffic} as well as travel time \cite{wu2004travel}. Compressed sensing is exploited by \cite{mitrovic2015low} to reduce the complexity of network, then support vector regression (SVR) is used for predicting travel speed on a Nationwide traffic network in Singapore. \cite{qi2014hidden} applied hidden Markov Model by incorporating traffic volume, lane occupancy, and traffic speed data in the model, using data from a 38 mile corridor of I-4 in Orlando, FL. In addition, \cite{ramezani2012estimation} and \cite{yeon2008travel} also used Markov chains to predict travel time on arterial routes. In those studies, oftentimes features used to train machines are limited to the road segment of its own. Very few spatio-temporal features were incorporated, which can drastically improve prediction performance, particularly under incidents.

Recently, studies have taken spatial-temporal correlations into consideration when predicting link travel time or flows. \cite{kamarianakis2003forecasting} considered spatial correlations as function of distance and degree of neighbors when applying a multivariate autoregressive moving-average model to the forecasting of traffic speed, which was tested on a dataset collected via 25 loop-detectors at Athens, Greece. \cite{kamarianakis2012real} discusses the extensions of time series prediction model such as ARIMA by considering correlations among neighbors and the utilization of LASSO for model selection. Patterns of the spatial and temporal prediction errors are inferenced through k-means clustering as well as PCA \cite{asif2014spatiotemporal}. \cite{salamanismanaging} defined a graph based Coefficient of Determination (CoD) matrix and utilized a modified BFS algorithm to reduce the time complexity of calculating the CoD matrix. On top of that, a graph based lag-STARIMA is proposed and used for travel time prediction. \cite{hunter2013arriving} proposes a method using temporal Bayesian network.  \cite{zou2014space} introduces a space–time diurnal (ST-D) method in which link-wise travel time correlation at multiple lag time is utilized. \cite{sun2012network} utilizes Gaussian process regression (GPR) model and graphic Lasso to forecast traffic flow. \cite{cai2016spatiotemporal}proposes a KNN model to forecast travel time up to one hour ahead, the model uses redefined inter-segments distances by incorporating the grade of connectivity between road segments, and considers spatial-temporal correlations and state matrices to identify traffic state.  \cite{min2011real} proposes a modified multivariate spatial-temporal autoregressive (MSTAR) model by leveraging the distance and average speed of road networks to reduce the number of parameters, the algorithm was tested on a road network of 502 links of which the traffic status are collected by loop detectors, the results remains accurate for up to one hour. 

Recent development in deep learning models in general has also accelerated the advancement of traffic prediction models. Several trials have been done using (deep) neural network to estimate short-term travel time, e.g., \cite{cui2019traffic}. In particular, \cite{ma2015large} proposes a restricted Boltzmann Machine (RBM)- based RNN model with two layers, the output are binary for each link: congested or not. Matrix are used to represent transportation network, no spatial information is utilized. The model is tested using taxi GPS data of Shenzhen, China. \cite{duan2016deep} proposes a deep learning method to impute traffic flow data by taking into consideration a few spatial and temporal factors, such as weather and day of week. \cite{polson2017deep,cui2020learning} exploit the spatio-temporal relations of network traffic have been widely used to make accurate and ahead-of-curve traffic predictions on a network level. 

\section{Dataset}
\label{sec: dataset}
\begin{figure}
    \centering
    \includegraphics[width=1\textwidth]{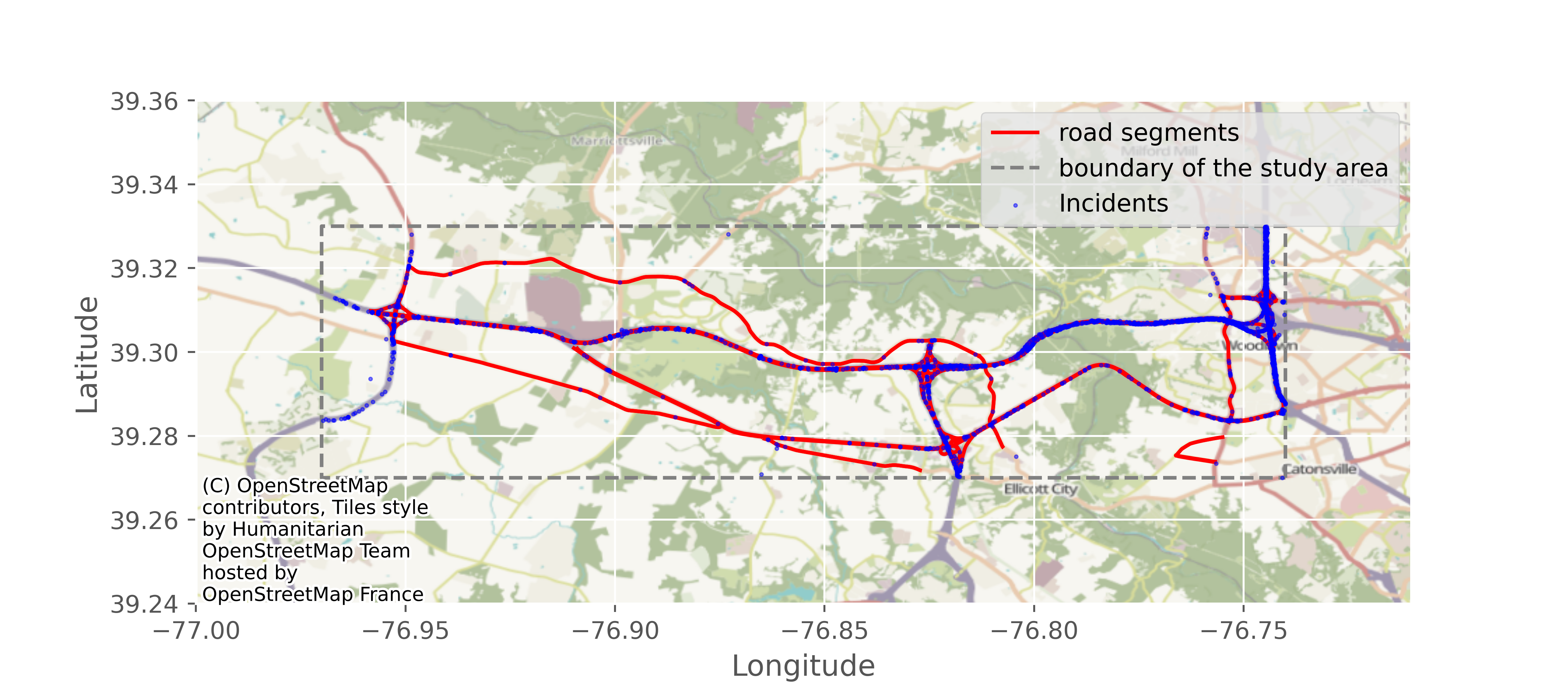}
    \caption{The TSMO network.}
    \label{fig: TSMO net}
\end{figure}

This paper examines a Transportation Systems Management and Operations (TSMO) network in Maryland US, as illustrated in Figure \ref{fig: TSMO net}. The TSMO network includes Interstate 70, US Route 29, and US Route 40, extending from the MD 32 interchange to Interstate 695. We utilize three primary data sources: (1) probe-sourced traffic speed measurements from INRIX\footnote{\url{https://inrix.com/products/ai-traffic/}}, (2) crowd-sourced traffic incident reports from Waze\footnote{\url{https://www.waze.com/}}, and (3) weather data from Weather Underground\footnote{\url{https://www.wunderground.com/}}. Historical datasets from INRIX and Waze are maintained by the Regional Integrated Transportation Information System (RITIS) platform\footnote{\url{https://ritis.org/}}, while Weather Underground data were retrieved using the Wunderground API.

The TSMO network comprises 181 link segments. The multi-sourced dataset was compiled by organizing different source data at a 5-minute frequency from 5:30 AM to 8:59 PM daily, covering the period from February 14, 2022, to February 13, 2023. Raw data with a frequency shorter than 5 minutes were aggregated to match the 5-minute frequency, while linear interpolation was applied to time series with missing entries or longer collection intervals. 

\subsection{INRIX traffic speed data}
The historical traffic speed measurements by INRIX were acquired from RITIS, encompassing major US highways and arterials. The INRIX data, collected at 5-minute intervals, relates to road segments identified by the INRIX Traffic Message Channel (TMC) code. Each record includes the TMC segment code, timestamp, observed speed (mph), average speed (mph), reference speed (mph), and two entries reflecting the confidence levels: the confidence score and confidence value. The dataset covers the period from February 14, 2022, to February 13, 2023. We fill in missing values by applying linear interpolation.

\subsection{Waze incident data}
Waze is a mobile navigation application, which not only includes navigation information but also event information such as traffic crashes, car breakdowns, congestion, hazards, and policy presence. The event information is provided and updated based on data provided by Waze users. Users can initiate new reports on events, or check whether the event is still present. The Waze data is queried from RITIS. The data span from February 14, 2022, to February 13, 2023, which are represented by dots in Figure \ref{fig: TSMO net}. 

\subsection{Weather underground data}
Weather Underground provides hourly weather measurements, detailing temperature, pressure, dew point, humidity, wind speed, hourly precipitation, visibility, and a text description of the overall weather conditions.

\section{Methodology}
\label{sec: method}
\subsection{Data pipeline for multi-source data}
In this study, multi-source data is utilized to predict the traffic speed of multiple road segments, which can be formulated as an univariate or multivariate problem. To predict effectively, the model should take as input normalized and aligned features. Therefore, this section provides a data pipeline to transform raw multi-source data into predictive features. First, the raw data is processed to be features. Then, the features are aligned both spatially and temporally.

\subsubsection{Data processing}
\textbf{Traffic speed} 
The raw speed data is organized from INRIX probe vehicle data in TMC (Traffic Message Channel) resolution with a temporal resolution of 5 minutes. The slowdown speed (SD) is then extracted from the organized TMC speed data by Equations (\ref{eqn: SD}). The reason of including SD is to provide the model with indicators of traffic congestion and incidents and denoise the incident features, which will be elaborated on later.

\begin{equation}
\text{SD}_{i}^{d,t} = \max(\frac{\sum_{j\in\Gamma^{-1}}y_{j}^{d,t}}{N_i}-y_{i}^{d,t}, 0) \label{eqn: SD}
\end{equation}
where $y_{i}^{d,t}$ denotes the speed of link $i$ at day $d$ time $t$, and fraction term computes the average speed of $N_i$ upstream links, the set of which is denoted by $\Gamma^{-1}$.\\

\textbf{Incident features} 
The incident status of targeted road segments is initially collected from Waze. Several types of incidents, including accidents, road-closing events, and hazard floods, are deemed critical and chosen for our study. Because Waze does not record all critical events or anomalies and some Waze reports have trivial impacts on traffic states, we propose an algorithm to denoise the Waze incident data, which is summarized in algorithm \ref{alg:1}. First, the top-$n$ percent of slowdown speed is identified as the abnormal slowdown, which identifies unreported incidents. Then, for each incident reported by Waze, if there is no abnormal slowdown during the reported incident interval, the incident is regarded as insignificant and removed. Subsequently, the value of $n$ is adjusted if the removal and adding percentages do not satisfy the thresholds. If all thresholds are met, the denoised incident indicator matrix is output after adding unreported anomalies and deleting insignificant incidents.

Given the denoised incident indicators, three incident features are produced, including segment incident indicator, network incident indicator, and incident count. The segment incident indicator reflects whether there are any incidents on this road segment at this moment, while the network incident indicator reflects whether there are any incidents on any road segments in the entire road network. The incident count is the number of incidents happening in the network at this time.

\begin{algorithm}
\begin{footnotesize}
\begin{algorithmic}
\caption{Incident data denoising}
\label{alg:1}
\State \underline{\textbf{Inputs}}: \parbox[t]{\dimexpr\linewidth-\algorithmicindent}{
\begin{itemize}
    \item incident report binary matrix $\mathbf{INC^i}\in \mathbb{R}^{\text{number\_of\_days} \times \text{time\_of\_the\_day}}$
    \item slowdown speed matrix $\mathbf{SD^i}\in \mathbb{R}^{\text{number\_of\_days} \times \text{time\_of\_the\_day}}$
    \item removal percentage threshold $\mathbf{\theta_1}$
    \item addition percentage threshold $\mathbf{\theta_2}$
    \item minimum duration for a prolonged anomaly to be labeled $\theta_t$
\end{itemize}
}
\State \underline{\textbf{Outputs}}: denoised incident feature matrix $\mathbf{DII^i}\in \mathbb{R}^{\text{number\_of\_days} \times \text{time\_of\_the\_day}}$
\State \underline{\textbf{Initialization}}: top $\mathbf{n}\%$ slowdown speed is the indicator of anomalies

\State \underline{\textbf{Step 1: Generate the abnormal slowdown speed matrix}}:
\State obtain the abnormal slowdown speed threshold:  $\mathbf{\theta_{SD}^i} = \text{Percentile}(\text{vec}(\mathbf{SD^i}), n)$
\State compute the abnormal slowdown speed binary matrix: $\mathbf{ASD^{i}_{(p,q)}} = \mathbb{I}(\mathbf{SD^{i}_{(p,q)}} \geq \mathbf{\theta_{SD}^i})$

\State \underline{\textbf{Step 2: Remove insignificant incident reports}}: 
\State initialize the significant incident report matrix $\mathbf{SIR^i}$ with zeros: $\mathbf{SIR^i} \gets \mathbf{0}^{\text{number\_of\_days} \times \text{time\_of\_the\_day}}$ 
\For {reported incident intervals $(p, q:q+r)$}
    \If{\text{SUM}($\mathbf{ASD^i_{(p,q:q+r)}}$) $\geq$ 1}
    \State $\mathbf{SIR^i_{(p,q:q+r)}} \gets 1$
    \EndIf
\EndFor

\State \underline{\textbf{Step 3: Check the removal and addition percentage threshold}}: 
\State compute the removal percentage $\mathbf{rm\%}$  = 1 - ($\text{SUM}(\mathbf{SIR^i}) / \text{SUM}(\mathbf{INC^i})$)
\State determine which labels are newly added $\mathbf{ADD^i} = \text{ReLU}(\mathbf{ASD^i} - \mathbf{SIR^i}, 0)$
\State compute the addition percentage $\mathbf{add\%}$ = $\text{SUM}(\mathbf{ADD^i}) / \text{SUM}(\mathbf{INC^i})$
\If{$\mathbf{rm\%} > \mathbf{\theta_1}$ and $\mathbf{add\%} \leq \mathbf{\theta_2}$}
\State $n \gets n+\alpha$
\State Go back to \textbf{Step 1}
\ElsIf{$\mathbf{rm\%} \leq \mathbf{\theta_1}$ and $\mathbf{add\%} > \mathbf{\theta_2}$}
\State $n \gets n-\alpha$
\State Go back to \textbf{Step 1}
\ElsIf{$\mathbf{rm\%} \leq \mathbf{\theta_1}$ and $\mathbf{add\%} \leq \mathbf{\theta_2}$}
\State Go to \textbf{Step 4}
\Else
\State the settings of $\mathbf{\theta_1}$ and $\mathbf{\theta_2}$ are unreasonable, terminate the algorithm and reinitialize
\EndIf

\State \underline{\textbf{Step 4: Generate denoised incident features}}: 
\State combine significant incident report and newly added labels by $\mathbf{DII^i} = \mathbf{SIR^i} + \mathbf{ADD^i}$ 
\State output $\mathbf{DII^i}$
\end{algorithmic}
\end{footnotesize}
\end{algorithm}

\textbf{Time features}
We consider four types of time variables, including hour-of-day, day-of-week, month-of-year, and official US holidays. For these time features, cyclic sine and cosine functions are used to transform them into numerical features $[t_i^{(\sin)}, t_i^{(\cos)}]$ as follows.

\begin{align}
t_i^{(\sin)} = \sin(2\pi i/T)\\
t_i^{(\cos)} = \cos(2\pi i/T)
\end{align}
where \(i\) denotes the hour or month index and \(T\) denotes the total number of hours in a day (\(T=24\)) or the number of months (\(T=12\)) in a year. A benefit of cyclic encoding is that time variables are mapped onto a circle, so the lowest values are next to the largest value (e.g. 0 am is next to 11 pm). Holidays are encoded into binary variables to indicate whether it is a holiday in the US.

\textbf{Weather features}
The weather information is extracted from the Weather Underground. The processing follows the procedures specified in \cite{Yao_2020}. All the processed features are summarized in Table \ref{tab: features}.

\begin{table}[]
\caption{Features}
\label{tab: features}
\centering
\begin{tabular}{l p{10cm}}

\hline
Feature name       & Explanation                                                                                     \\
\hline

\multicolumn{2}{c}{\textbf{Incident features}}                                                                                \\
Segment incident indicator & Binary variable. 1 if there are any incidents in the road segment, else 0.   \\

Network incident indicator & Binary variable. 1 if there are any incidents in the network, else 0.                                \\

Incident count     & Numerical variable. The number of incidents in the network.                                 \\
\multicolumn{2}{c}{\textbf{Time features}}                                                                                    \\
Time of day        & Numerical variable. Cyclic encoding of hour of day.                                             \\
Month of year      & Numerical variable. Cyclic encoding of month of year.                                           \\
Day of week        & Numerical variable. Cyclic encoding of day of week.                                                 \\
Holiday indicator  & Binary variable. 1 if it is a holiday, else 0.\\           
\multicolumn{2}{c}{\textbf{Weather features}}                                                                                 \\
Temperature        & Numerical variable. Air temperature in degree F.                                                \\
Dew point          & Numerical variable. Target air temperature in degree F to achieve a relative humidity of 100\%. \\
Humidity           & Numerical variable. Relative humidity in percent.                                               \\
Wind speed         & Numerical variable. Wind speed in mile per hour.                                                \\
Wind gust          & Numerical variable. Gust wind speed in mile per hour.                                           \\
Precipitation      & Numerical variable. Precipitation in inches.                                                    \\
Weather condition  & Categorical variable. The description of the weather condition. E.g., 'Cloudy' and 'Fair'.      \\
\hline
\end{tabular}
\end{table}

\subsubsection{Data integration}
Data in this study are collected from multiple sources, so they initially are not aligned either in spatial or temporal. To leveraging these data to predict traffic state, the following spatial and temporal integration steps are conducted.

\textbf{Spatial integration} INRIX speed data is spatially labeled by TMC road segment code, so TMC road segments serve as the spatial integration basis in our study. Waze incidents are linked to the TMC road segments based on the incident locations. The weather data represents the weather conditions of the entire network area, so the weather features are identical along different road segments at same time.

\textbf{Temporal integration}. Different data sources collect data in different time intervals. Waze events are reported by Waze users in real-time without specific time intervals, and the Underground Weather data is hourly measurements. INRIX speed data is collected at 5-minute intervals, so we map all features into 5-minute intervals. Waze data is interpolated or aggregated into 5-minute intervals, and we resample hourly Weather data and fill in the missing values.

\subsection{Model structure}

\begin{figure}
    \centering
    \includegraphics[width=0.9\textwidth]{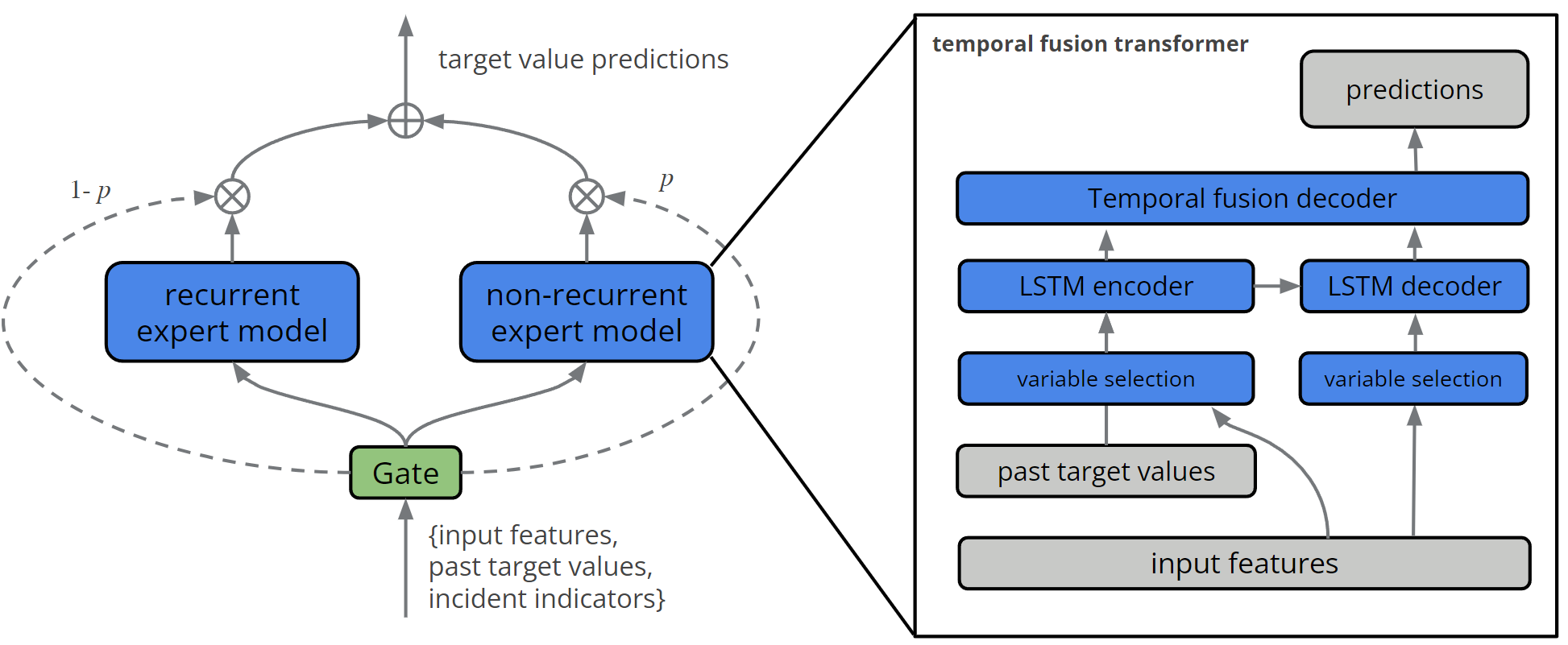}
    \caption{The model overview.}
    \label{fig: model}
\end{figure}

\subsubsection{Mixture of experts}
The Mixture of Experts (MoE) model is a machine learning approach that combines the strengths of multiple specialized models to solve complex prediction tasks. Each "expert" within the mixture is typically a separate neural network or other machine learning models. A gate or router dynamically assigns input data to the most appropriate experts. This gate determines the weighting of each expert's output based on the input features. The MoE approach is advantageous for predicting recurrent and non-recurrent traffic states due to its ability to handle diverse and complex patterns in traffic data. By employing multiple specialized models (experts) and a gate to dynamically allocate input data, MoE allows each expert to focus on specific aspects of traffic patterns, leading to more accurate and robust predictions. Besides, this method enhances model interpretability, as it highlights the factors influencing different traffic states, and offers flexibility and scalability, allowing for the addition of new experts without retraining the entire model.

We adopt the MoE model structure to suit the diverse needs of predictions in recurrent and non-recurrent conditions. Specifically, our model is composed of independent recurrent and non-recurrent components, thereby enabling efficient adaptation to diverse traffic conditions.
\begin{multline}
    \hat{\yv}_t = \underbrace{(1 - p_t) \odot \boldsymbol{\mu}_\phi (\Xv_{t-c},\cdots,\Xv_{t}, \cdots, \Xv_{t+h-1}; \yv_{t-c}, \cdots, \yv_{t-1})}_\text{recurrent component} \\
    + \underbrace{p_t \odot \boldsymbol{\mu}_\theta (\Xv_{t-c},\cdots,\Xv_{t}, \cdots, \Xv_{t+h-1}; \yv_{t-c}, \cdots, \yv_{t-1})}_\text{non-recurrent component}
\end{multline}
where $\hat{\yv}_t$ is predicted link speeds at time $t$, and $\yv$ is the past speed values. $\boldsymbol{\mu}_\phi (\cdot)$ is the recurrent expert model, and $\boldsymbol{\mu}_\theta (\cdot)$ is the non-recurrent expert model. The input of expert models includes input features $\Xv$ from the beginning of context window $t-c$ to the end of prediction window $t+h-1$ and past speed values from $t-c$ to $t-1$. $p_t$ is the assigned weight to the output of the non-recurrent expert model. Our model overview is plotted in Figure \ref{fig: model}.

\subsubsection{Expert model}
Though the choice of expert models is flexible, the temporal fusion transformer (TFT) model \cite{lim2021temporal} is adopted as the expert model backbone in our study for two reasons. First, TFT demonstrated superior time series forecasting performances in various types of datasets including traffic and electricity \cite{lim2021temporal}. More importantly, unlike other deep learning models that work as black boxes, TFT is more interpretable by capturing the temporal dependencies and quantifying the global feature importances.

The main blocks within TFT are plotted on the right of Figure \ref{fig: model}. The input features and past target values first go through variable selection blocks, in which gated residual networks are utilized to assign variable selection weights to all input variables. Then, the weighted input features in the context time window and the weighted past target values are sent to a long short-term memory (LSTM)-based encoder, while the weighted input features in the prediction window are sent to an LSTM-based decoder. Subsequently, a temporal fusion decoder calculates the multi-head attention among embeddings of different time steps similar to the attention mechanism in the transformer model \cite{vaswani2017attention}. These blocks enable TFT to learn the temporal dependencies and the variable importance, which bring not only strong forecasting performances but also superior interpretability. For details about TFT, we refer to \cite{lim2021temporal}.

\subsection{Model training}
To account for temporal data distribution shift issues, the dataset was divided into training (80\%), validation (10\%), and test (10\%) sets over time. Therefore, the least recent data is used for model training, and models demonstrating the best performance on the validation set are selected to test on the most recent data. 

In time series forecasting problems, there are typically two time windows, including a context window of length $c$ and a prediction window of length $h$. The context window is how far the model looks back, and the prediction window is how far the model predicts. Thus, the datasets are processed into data slices of length $c+h$ using a sliding window with a step size of 1.

\begin{figure}
    \centering
    \includegraphics[width=0.8\textwidth]{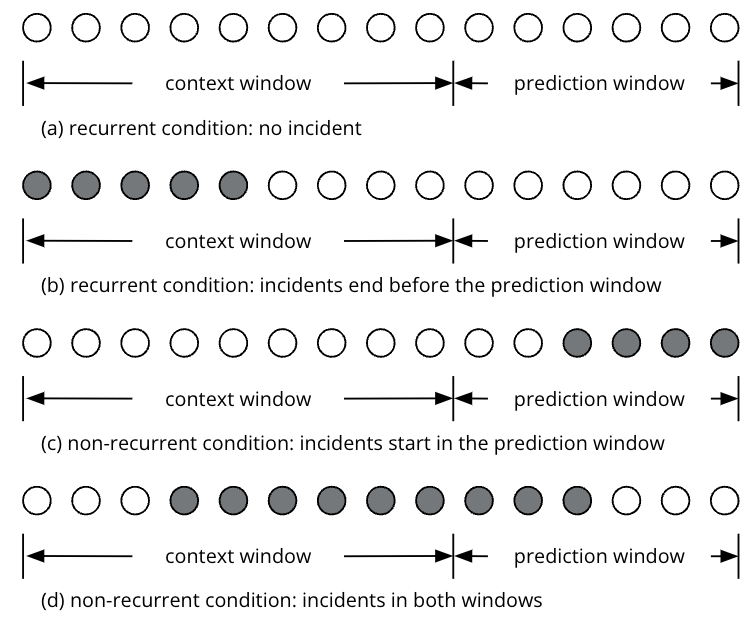}
    \caption{The definition of recurrent and non-recurrent conditions.}
    \label{fig: NR definition}
\end{figure}

Given the observation that traffic patterns with occurrences of incidents deviate from recurrent traffic patterns, we define recurrent and non-recurrent conditions based on the incident occurrences within the road network. Figure \ref{fig: NR definition} demonstrates the four most common conditions. The solid circles in Figure \ref{fig: NR definition} are time steps when there are any incidents within the road network, while the hollow circles are time steps without any incidents. If all time steps in both windows are incident-free, it is obviously a recurrent condition. If an incident happens in the context window but ends before the prediction window, it is categorized as a recurrent condition. If the incident happens in the prediction window or the incident happens in the context window but lasts to the prediction window, it is regarded as a non-recurrent condition.

The data distribution of the non-recurrent conditions is different from recurrent conditions because of the occurrences of incidents. Therefore, the recurrent expert model is trained on the train data in recurrent conditions only. For the non-recurrent expert model, we first pre-train it on the train data in recurrent conditions, and then fine-tune it on the train data in non-recurrent conditions. This training procedure for the non-recurrent model is designed for two reasons. First, the non-recurrent conditions are not as prevalent as the recurrent conditions, so the dataset size of non-recurrent conditions is not relatively small. Thus, using non-recurrent data only is not desirable for deep learning models that typically desire large datasets. Besides, we think recurrent data contain some common and general traffic patterns that may be helpful for prediction in non-recurrent conditions. In the above train, pre-train, and fine-tune processes, the quantile losses in \cite{lim2021temporal} are adopted. 

\section{Experiments and results}
\label{sec: experiments}
\subsection{Experiment setup}

Aside from the TFT model, this study includes the following baselines for comparison.
\begin{itemize}
    \item \textbf{LOb (last observation)} In LOb, the latest observed speed in the context window on the road segment is used to predict the speed in the prediction window,
    \begin{equation}
        \hat{\yv}_{t:t+h-1} = \yv_{t-1}
    \end{equation}
    \item \textbf{DeepAR (deep autoregressive model \cite{salinas2020deepar})} DeepAR utilizes an LSTM-based recurrent neural network to learn the probabilistic distributions of multiple related time series \cite{salinas2020deepar}. During inference, DeepAR outputs probabilistic distributions, from which the predictions are sampled. Though DeepAR is typically trained on multiple time series, it is univariate time series forecasting.
    \item \textbf{MVAR (multivariate autoregressive \cite{salinas2019high})} Similar to DeepAR, MVAR adopts an LSTM-based recurrent neural network. However, MVAR regards multiple time series as a multivariate probabilistic distribution, and a low-rank covariance structure is leveraged to model the high-dimensional covariate matrix. Predictions are sampled from the modeled multivariate probabilistic distribution during inference.
\end{itemize}

To evaluate the model performances, we calculate the Symmetric Mean Absolute Percentage Error (SMAPE) in equation \eqref{SMAPE} and Root Mean Squared Error (RMSE) in equation \eqref{RMSE} of traffic speed predictions of all prediction horizons and all road segments in the test dataset.
\begin{equation}
\label{SMAPE}
    \text{SMAPE} = \frac{1}{T} \sum_{t=1}^T \frac{1}{H} \sum_{h=1}^H \frac{1}{N} \sum_{n=1}^N 
    \frac{|\hat{y}_{n, t + h} - y_{n, t + h}|}{(|\hat{y}_{n, t + h}| + |y_{n, t + h}|) / 2} \times 100
\end{equation}

\begin{equation}
\label{RMSE}
    \text{RMSE} = \frac{1}{T} \sum_{t=1}^T \frac{1}{H} \sum_{h=1}^H \sqrt{ \frac{1}{N} \sum_{n=1}^N ( \hat{y}_{n, t + h} - y_{n, t + h})^2}
\end{equation}

\subsection{Model performances}

\begin{table}[]
\caption{Prediction errors of expert models on the recurrent/non-recurrent test dataset.}
\label{tab: expert models}
\centering
\begin{tabular}{@{}llll@{}}
\toprule
Method & SMAPE & RMSE \\ \midrule
\multicolumn{3}{c}{Recurrent} \\
LOb &  14.30 &  7.30\\
DeepAR-R &  11.42 & 5.58 \\
MVAR-R &  16.23 & 8.27 \\
TFT-R &  \textbf{10.95} & \textbf{5.46} \\
\multicolumn{3}{c}{Non-recurrent} \\
LOb & 15.13 & 7.61 \\
DeepAR-NR &  12.25  & 5.95 \\
MVAR-NR &  16.23  & 8.03 \\
TFT-NR &  11.75  & 5.79 \\
TFT-FT &  \textbf{11.65}  & \textbf{5.78} \\ \bottomrule
\end{tabular}
\end{table}

\begin{table}[]
\caption{Prediction errors on the entire test dataset.}
\label{tab: MoE models}
\centering
\begin{tabular}{@{}llll@{}}
\toprule
Method & SMAPE & RMSE \\ \midrule
LOb &  14.55 & 7.39 \\
DeepAR-ALL  & 11.85 & 5.78 \\ 
MVAR-ALL & 12.91 & 6.29 \\
TFT-ALL & 11.61 & 5.66  \\ 
MoE &  \textbf{11.16}  & \textbf{5.56} \\ \bottomrule
\end{tabular}
\end{table}

\begin{figure}[ht]
    \centering
    \subfigure[SMAPE]{
        \includegraphics[width=0.475\textwidth]{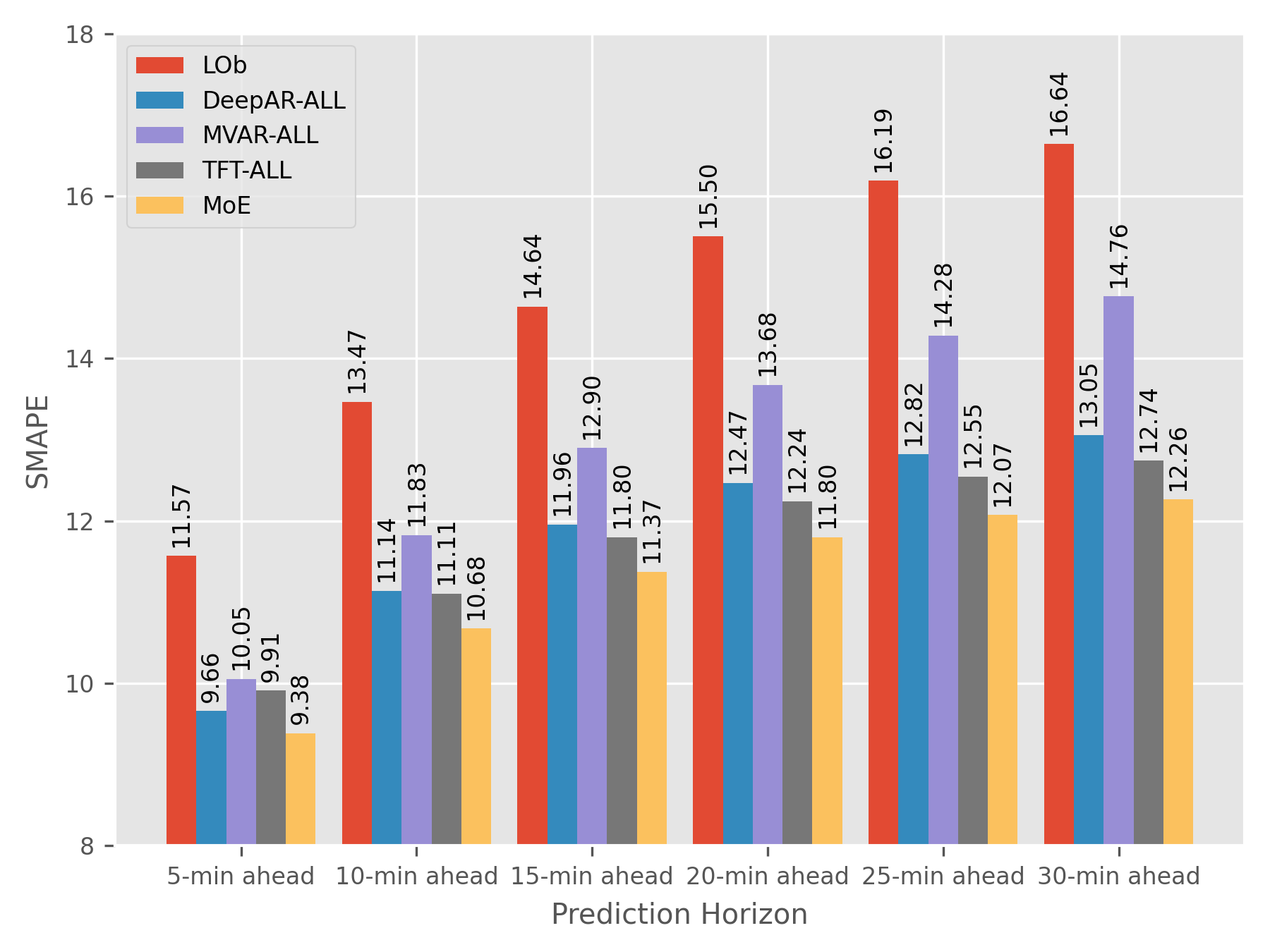}
    }
    \subfigure[RMSE]{
        \includegraphics[width=0.475\textwidth]{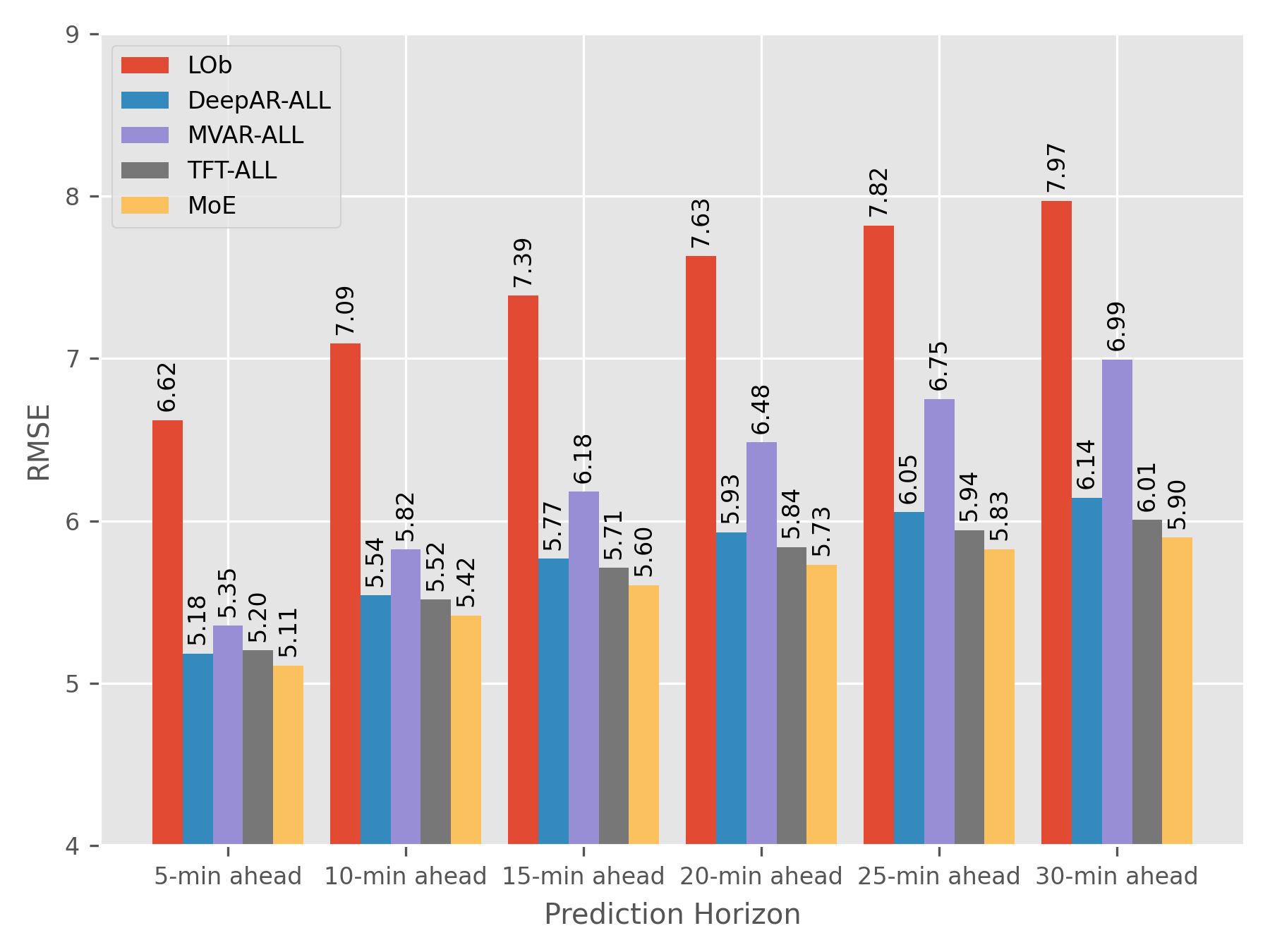}
    }
    \caption{Prediction errors along different prediction horizons on the entire test dataset.}
    \label{fig: stepwise errors}
\end{figure}

\begin{figure}[ht]
    \centering
    \subfigure[SMAPE]{
        \includegraphics[width=0.475\textwidth]{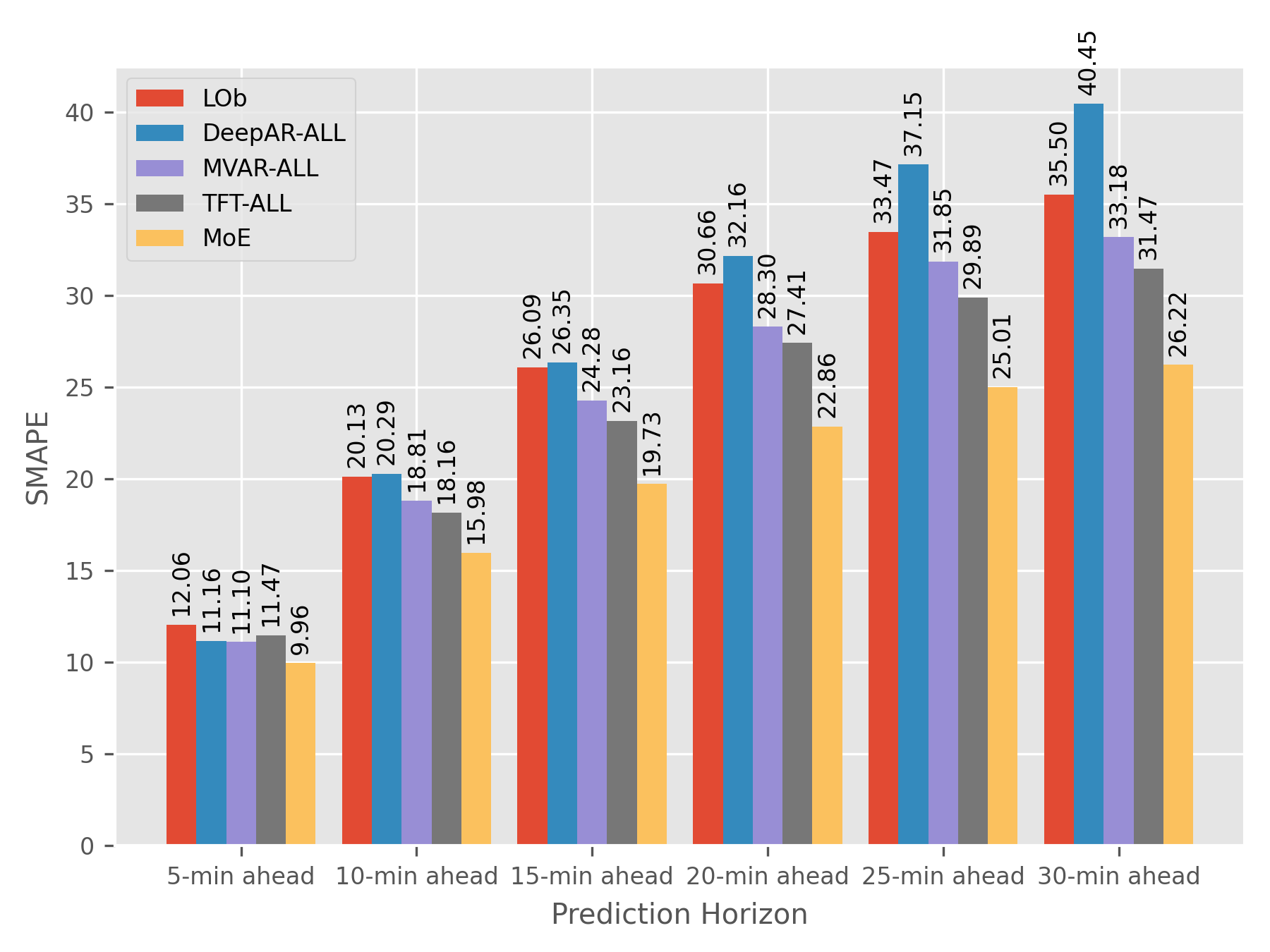}
    }
    \subfigure[RMSE]{
        \includegraphics[width=0.475\textwidth]{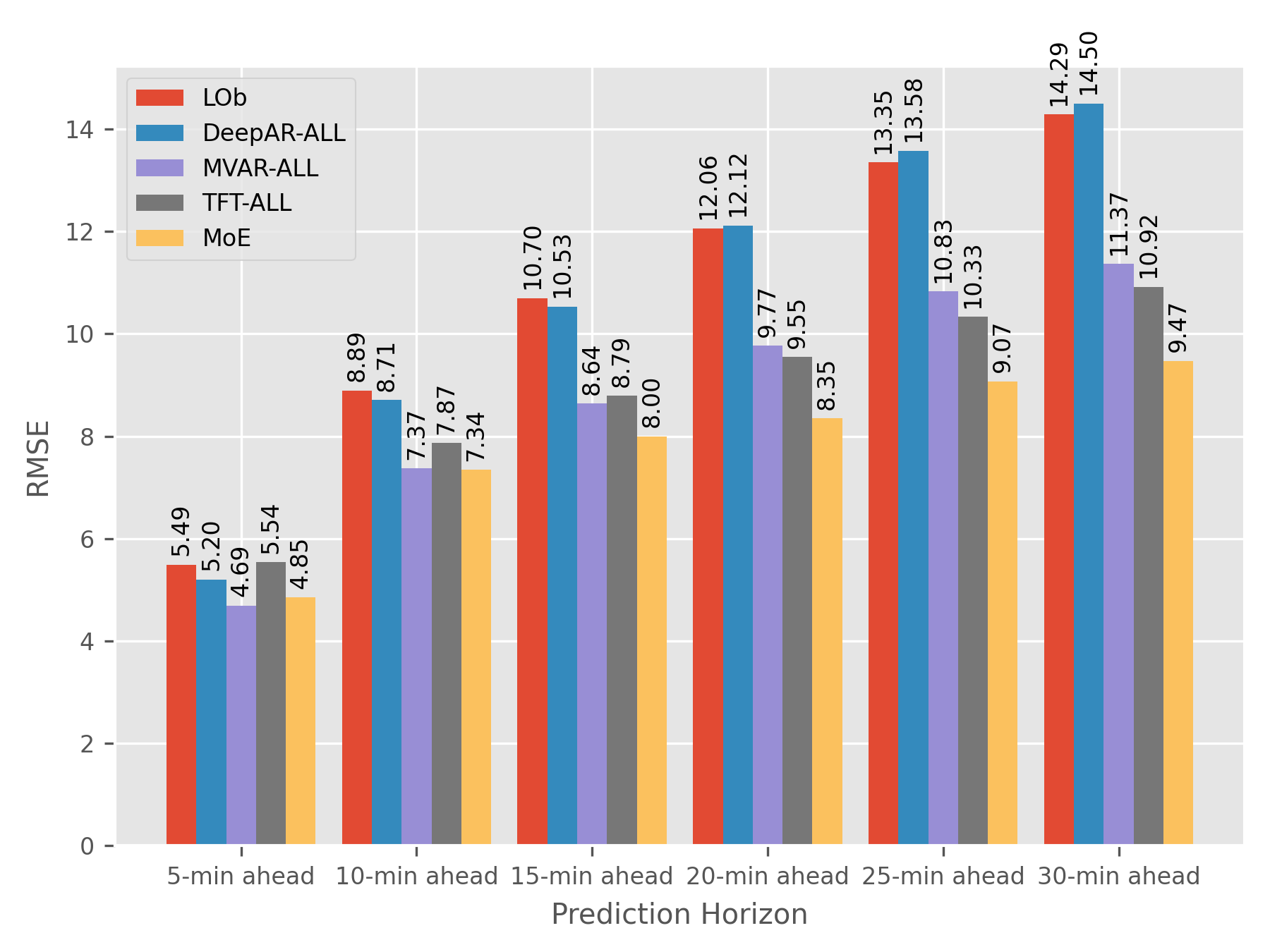}
    }
    \caption{Prediction errors along different prediction horizons on the incident road segments.}
    \label{fig: stepwise errors on incident links}
\end{figure}

\begin{figure}[ht]
    \centering
    \subfigure[TFT-ALL on example 1]{
        \includegraphics[width=0.45\textwidth]{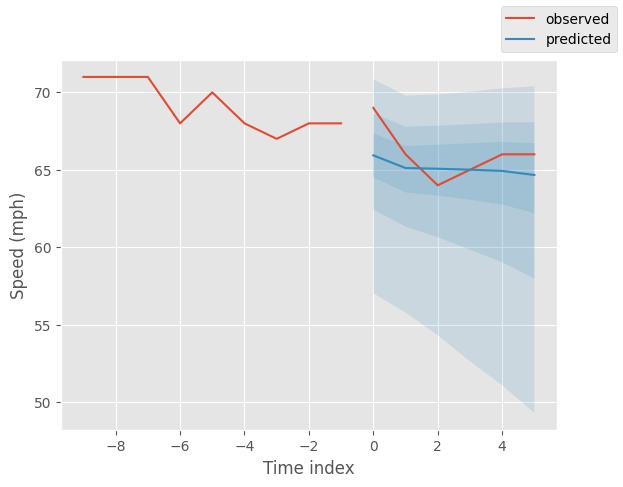}
    }
    \subfigure[MoE on example 1]{
        \includegraphics[width=0.45\textwidth]{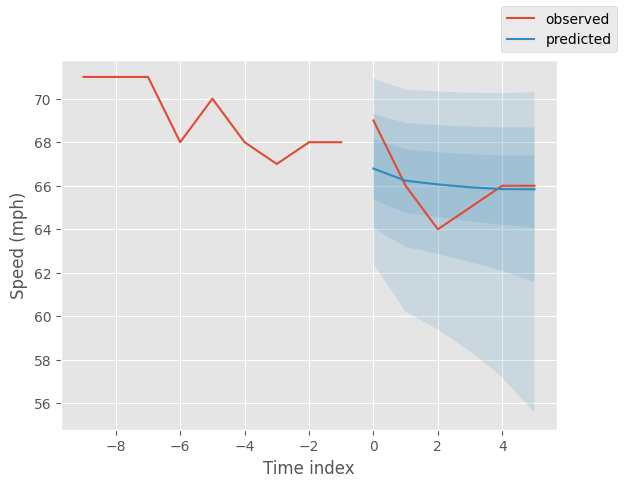}
    }
    \subfigure[TFT-ALL on example 2]{
        \includegraphics[width=0.45\textwidth]{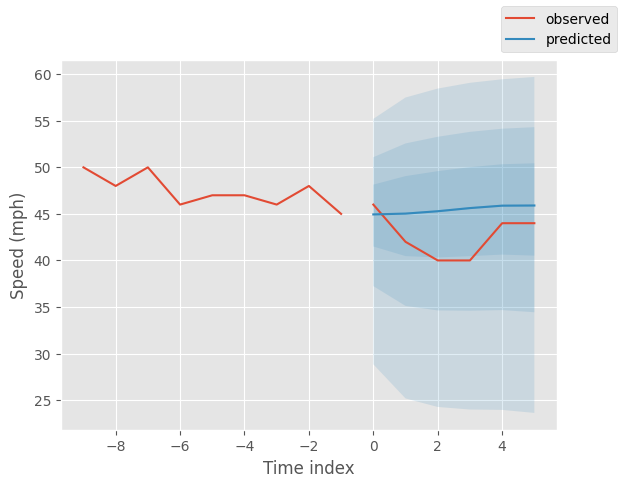}
    }
    \subfigure[MoE on example 2]{
        \includegraphics[width=0.45\textwidth]{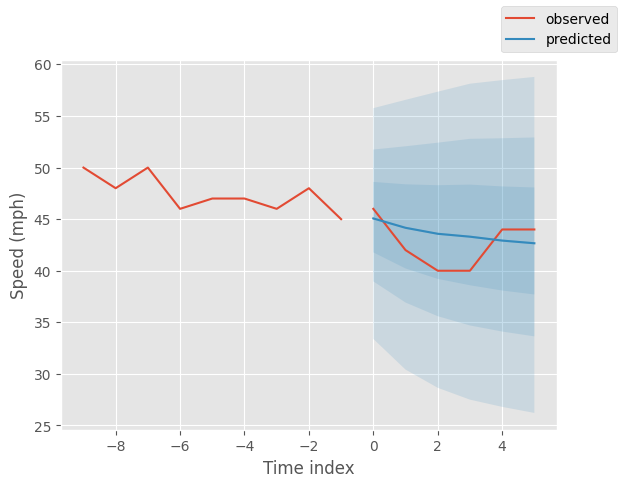}
    }
    \caption{Predictions of MoE and TFT-ALL on non-recurrent examples.}
    \label{fig: examples}
\end{figure}

First, the expert models are evaluated on the recurrent and non-recurrent test datasets respectively. Table \ref{tab: expert models} summarized the evaluation results. DeepAR, MVAR-R, and TFT-R are models trained on the recurrent dataset, while DeepAR-NR, MVAR-NR, and TFT-NR are trained on the non-recurrent dataset. Note that TFT-FT is a TFT model that is firstly pre-trained on the recurrent dataset and then fine-tuned on the non-recurrent dataset. 

The results of recurrent expert models show that TFT-R leads to the lowest SMAPE and RMSE. Besides, MVAR-R is worse than DeepAR-R, indicating that formulating the traffic speed prediction as a multivariate time series forecasting problem is not desirable in our study. One possible reason is that the efforts to model the correlation between time series hinder the model training efficiency.

The results of non-recurrent expert models also show the superiority of TFT models. TFT-NR and TFT-FT are the two best models. Notably, TFT-FT performs better than TFT-NR, which validates the efficacy of our proposed training pipeline for non-recurrent models. 

Comparing the results of recurrent and non-recurrent, SMAPE and RMSE in non-recurrent conditions are generally higher than in recurrent conditions. This observation is expected, because traffic patterns in the non-recurrent conditions are less predictable due to the disruptive impacts of incidents.

Then, the MoE model is tested on the entire test dataset. For comparison, we also train DeepAR, MVAR, and TFT models on the entire train dataset, which are denoted as DeepAR-ALL, MVAR-ALL, and TFT-ALL respectively. The results are summarized in Table \ref{tab: MoE models}. Among all baselines, TFT-ALL is the best, and our proposed MoE is better than the TFT-ALL, which proves the functionality of the proposed MoE structure.

In time series forecasting problems, how the prediction errors propagate along prediction horizons is of interest. We first show the prediction errors on the entire test dataset in Figure \ref{fig: stepwise errors}. Among all different prediction horizons, MoE leads to the lowest SMAPE and RMSE. Besides, both types of errors increase along prediction horizons as expected, but the increased magnitude of MoE is the smallest.

In non-recurrent conditions, the prediction performances on the road segments where incidents happen are critical for proactive management to mitigate incident impacts. To this end, the models are evaluated on the incident road segments during incident occurrences, as Figure \ref{fig: stepwise errors on incident links} shows. Interestingly, DeepAR-ALL which performs well in the entire test dataset, is sometimes even worse than LOb in this test. This observation indicates that DeepAR-ALL learns to fit the main recurrent pattern and regards these incident conditions as noise. Among all models, MoE leads to the lowest SMAPE and RMSE, and the gap between MoE and other baselines is more significant than the results on the entire test dataset.

To demonstrate how MoE outperforms TFT-ALL, we give two prediction examples of TFT-ALL and MoE in the non-recurrent conditions, as Figure \ref{fig: examples} show. In example 1, TFT-ALL and MoE both can capture the speed drop tendency caused by incidents, and predict that the speed continues to decline in the prediction horizon. However, in example 2 where the speed drop is less observable in the context window, TFT-ALL forecasts that the speed will be flat, while MoE successfully foresees the speed decrease. These examples specifically show MoE can adapt to non-recurrent conditions better than TFT-ALL.

\subsection{Interpretation}

\begin{figure}
    \centering
    \includegraphics[width=0.7\textwidth]{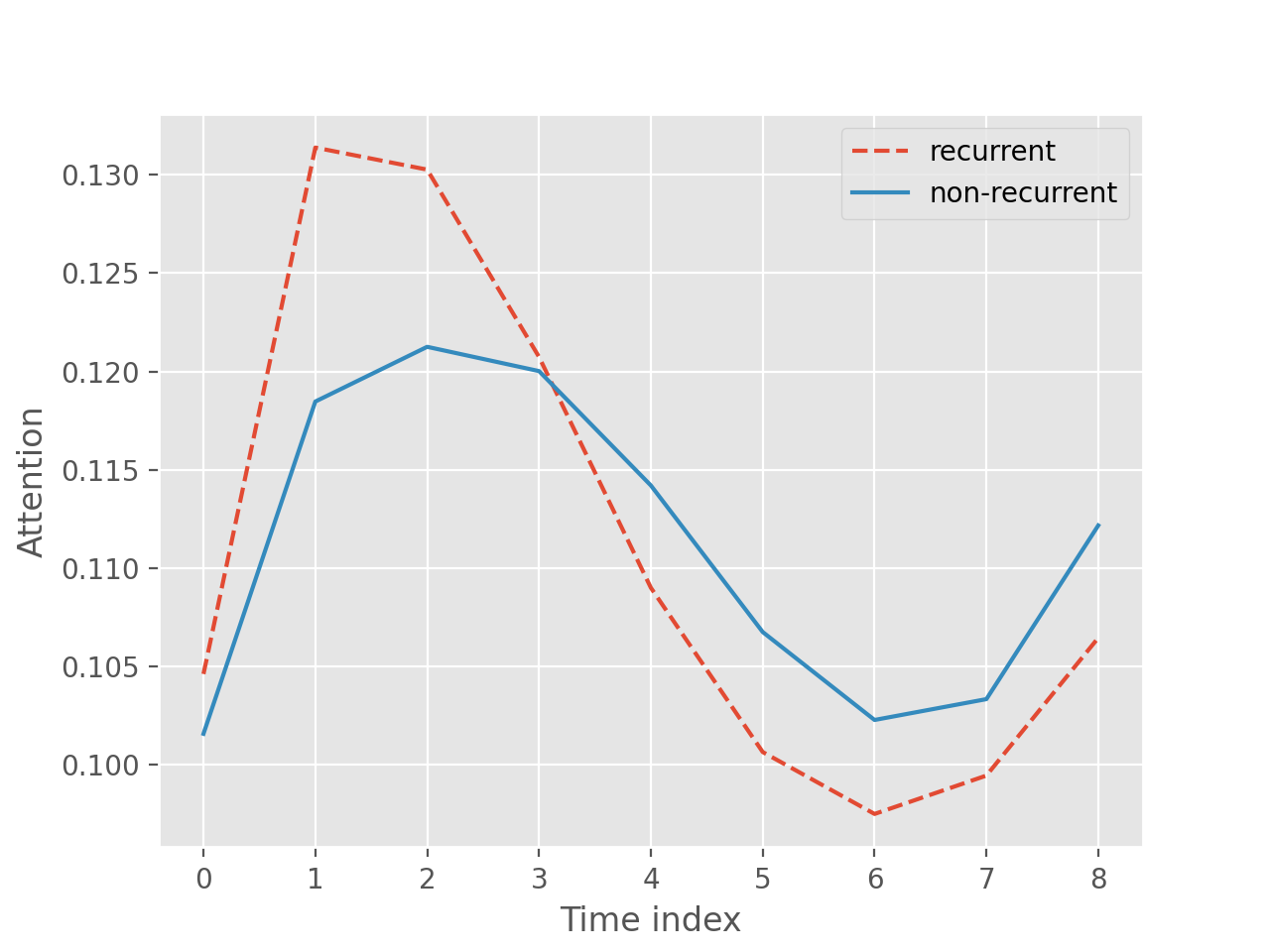}
    \caption{Temporal dependencies (attention) learned by MoE on test data.}
    \label{fig: attention}
\end{figure}

\begin{figure}[ht]
    \centering
    \subfigure[Encoder variables in the recurrent expert model.]{
        \includegraphics[width=0.45\textwidth]{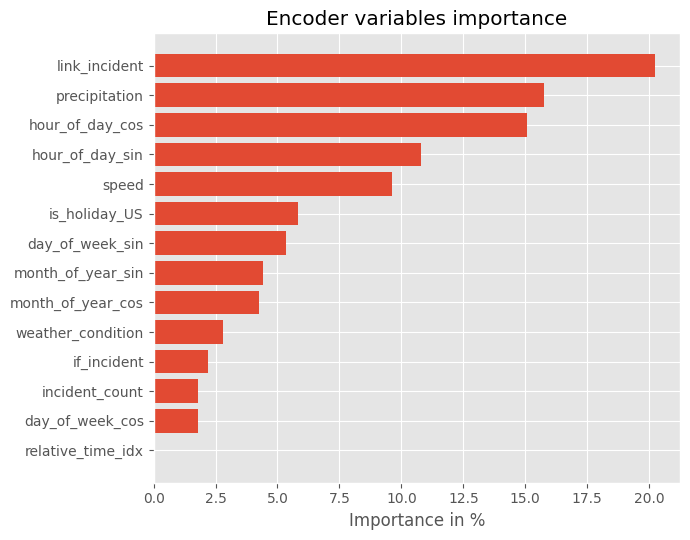}
    }
    \subfigure[Encoder variables in the non-recurrent expert model.]{
        \includegraphics[width=0.45\textwidth]{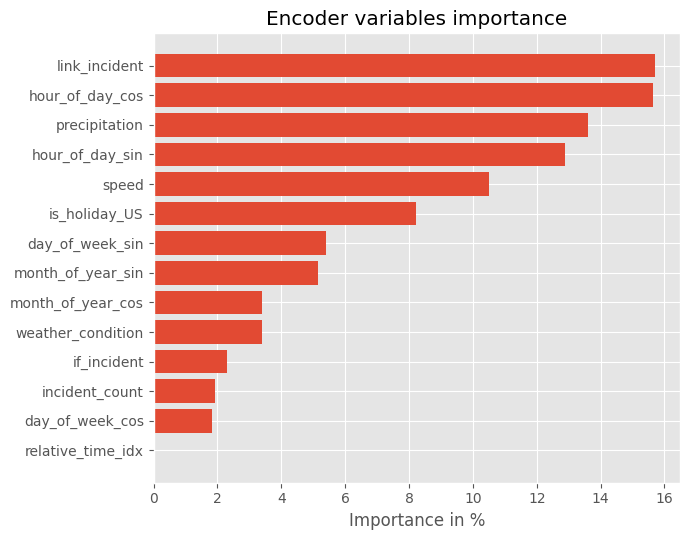}
    }
    \subfigure[Decoder variables in the recurrent expert model.]{
        \includegraphics[width=0.45\textwidth]{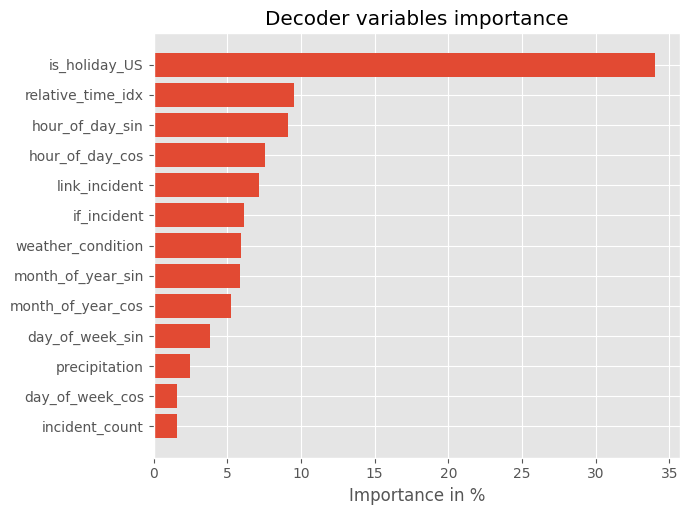}
    }
    \subfigure[Decoder variables in the non-recurrent expert model.]{
        \includegraphics[width=0.45\textwidth]{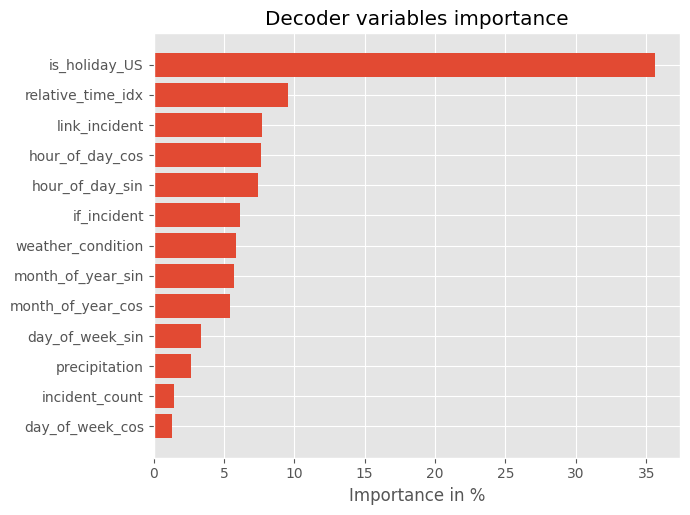}
    }
    \caption{Variable importance of recurrent and non-recurrent expert models.}
    \label{fig: variable importances}
\end{figure}

One research question of this study is trying to answer is how traffic prediction in non-recurrent conditions differs from recurrent conditions. Interpretation of patterns learned by our proposed MoE model may give some insights.

 In the context of deep learning, attention measures how much predictions depend on different vectors. In our study, attention quantifies the temporal dependencies between predicted speed and the values in the context window. The temporal dependencies learned by MoE are plotted separately for recurrent and non-recurrent conditions, as Figure \ref{fig: attention} shows. In both conditions, the largest dependencies are located at the beginning and the end of the context window, indicating the model learns the tendency of past values, such as speed drop and speed recovery. Compared with recurrent conditions, non-recurrent conditions depend more on the most recent time indices. Speed in non-recurrent conditions is volatile due to occurrences of incidents, so the predictions depend more on recent values.

Our models are trained on multi-source features, so we analyze the importance of various features. The variable importances are split by recurrent and non-recurrent conditions. The encoder variables include multi-source features and past speed values, while the decoder variables only contain the multi-source features. Importance percentages of the most important variables are plotted in Figure \ref{fig: variable importances}. Generally, we find 'speed' is not the most important feature, while 'link incident', 'precipitation', and 'hour of the day' are the most important. This indicates traffic speed either is impacted by incidents or follows a periodic pattern. The decoder variable importance shows holidays significantly affect traffic speed as well. Comparing the decoder variable importance in the recurrent conditions with the non-recurrent conditions, we can find the 'link incident' feature is more important in the non-recurrent conditions.



\section{Conclusion}
\label{sec: conclusion}
In conclusion, this study presents a Mixture of Experts (MoE) model that leverages recurrent and non-recurrent expert models to predict traffic states up to 30 minutes in advance. The MoE architecture allows the model to handle the diverse patterns of recurrent and non-recurrent traffic conditions, leading to superior prediction accuracy compared to baseline models. We also propose a novel training pipeline for non-recurrent models to accommodate limited data issues.

By interpreting the model prediction in recurrent and non-recurrent conditions separately, we shed light on the differences among two types of conditions. Future work could explore incorporating additional data sources, such as real-time incident severity information, to further enhance prediction accuracy.

\section{Acknowledgement}
This research is supported by US Department of Transportation Exploratory Advanced Research Award 693JJ321C000013. The contents of this paper reflect the views of the authors only, who are responsible for the facts and the accuracy of the information presented herein.

\bibliographystyle{plain} 
\bibliography{ref}

\end{document}